\documentclass[runningheads]{llncs}
\usepackage[dvipsnames]{xcolor}
\usepackage{ulem}
\usepackage{graphicx}
\usepackage{amsmath}
\usepackage{amsfonts}
\usepackage{multirow}
\usepackage{float}
\usepackage{bm}
\usepackage[colorlinks]{hyperref}
\usepackage{subcaption}
\usepackage{threeparttable}
\usepackage{subcaption}

\begin{document}
\title{FairAdaBN: Mitigating unfairness with adaptive batch normalization and its application to dermatological disease classification}
\titlerunning{FairAdaBN}
%
\author{Zikang Xu$^{1,2}$, Shang Zhao$^{1,2}$, Quan Quan$^3$, Qingsong Yao$^3$, and S. Kevin Zhou$^{1,2,3}$}

\authorrunning{Xu et al.}
%

\institute{School of Biomedical Engineering, Division of Life Sciences and Medicine, University of Science and Technology of China, Hefei, Anhui, 230026, P.R.China \and Suzhou Institute for Advanced Research, University of Science and Technology of China, Suzhou, Jiangsu, 215123, P.R.China \and
Key Lab of Intelligent Information Processing of Chinese Academy of Sciences (CAS), Institute of Computing Technology, CAS, Beijing 100190, China}
\maketitle              
\begin{abstract}
Deep learning is becoming increasingly ubiquitous in medical research and applications while involving sensitive information and even critical diagnosis decisions. 
Researchers observe a significant performance disparity among subgroups with different demographic attributes, which is called \textbf{model unfairness}, and put lots of effort into carefully designing elegant architectures to address unfairness, which poses heavy training burden, brings poor generalization, and reveals the trade-off between model performance and fairness. To tackle these issues, we propose \textbf{FairAdaBN} by making batch normalization adaptive to sensitive attributes. This simple but effective design can be adapted to several classification backbones that are originally unaware of fairness.
Additionally, we derive a novel loss function that restrains statistical parity between subgroups on mini-batches, encouraging the model to converge with considerable fairness.
In order to evaluate the trade-off between model performance and fairness, we propose a new metric, named Fairness-Accuracy Trade-off Efficiency (FATE), to compute normalized fairness improvement over accuracy drop.
Experiments on two dermatological datasets show that our proposed method outperforms other methods on fairness criteria and FATE. Our code is available at \url{https://github.com/XuZikang/FairAdaBN}.

\end{abstract}
\section{Introduction}

The past years have witnessed a rapid growth of applying deep learning methods in medical imaging
~\cite{zhou2021review}. 
As the performance improves continuously, researchers also find that deep learning models attempt to distinguish illness by using features that are related to a sample's demographic attributes, especially sensitive ones, such as skin tone or gender. 
The biased performance due to sensitive attributes within different subgroups is defined as \textbf{unfairness}~\cite{narayanan2018translation}.
For example, Seyyed-Kalantari~\textit{et. al.}~\cite{seyyed2020chexclusion} find that their models trained on chest X-Ray dataset show a significant disparity of True Positive Ratio (TPR) between male and female subgroups. Similar evaluations are done on brain MRI~\cite{petersen2022feature}, dermatology~\cite{kinyanjui2020fairness}, and mammography~\cite{lu2021evaluating}, which shows that unfairness issues exist extensively in medical applications.
If the unfairness of deep learning models is not handled properly, healthcare disparity increases, and human fundamental rights are not guaranteed. 
Thus, there is a pressing need on investigating unfairness mitigation to eliminate critical biased inference in deep learning models.

There are two groups of methods to tackle unfairness. The first group proceeds \uline{implicitly} with \textit{fairness through unawareness}~\cite{dwork2012fairness} by leaving out sensitive attributes when training a single model or deriving invariant representation and ignoring them subjectively when making a decision. However, plenty of evaluations prove that this may lead to unfairness, due to the entangled correlation between sensitive attributes and other variables in the data, and \textit{statistical difference} between features of different subgroups. The second group \uline{explicitly} takes sensitive attributes into consideration when training models, for example, train independent models for unfairness mitigation~\cite{puyol2021fairness,wang2020mitigating} with no parameters shared between subgroups. 
However, this may result in degraded performance because the amount of data for model building is reduced (see Table~\ref{tab:exp_fitz}). 

It is natural to consider whether it is possible to inherit the advantages from both worlds, that is, learning a single model on the whole dataset yet still with explicit modeling of sensitive attributes.
Therefore, we propose a framework with a powerful adapter termed \textbf{Fair Adaptive Batch Normalization (FairAdaBN)}.
Specifically, FairAdaBN is designed to mitigate task disparity between subgroups captured by the neural network. It integrates the common information of different subgroups dynamically by sharing part of network parameters, and enables the differential expression of feature maps for different subgroups, by adding only a few parameters compared with backbones. 
Thanks to FairAdaBN, the proposed architecture can minimize statistical differences between subgroups and learn subgroup-specific features for unfairness mitigation, which improves model fairness and reserves model precision at the same time. In addition, to intensify the models' ability for balancing performance and fairness, a new loss function named \textbf{Statistical Disparity Loss} ($L_{SD}$), is introduced to optimize the statistical disparity in mini-batches and specify fairness constraints on network optimization. $L_{SD}$ also enhances information transmission between subgroups, which is rare for independent models.
Finally, a perfect model should have both higher precision and fairness compared to current well-fitted models. However, most of the existing unfairness mitigation methods sacrifice overall performance for building a fairer model~\cite{sarhan2020fairness,suriyakumar2021chasing}. 
Therefore, following the idea of discovering the fairness-accuracy Pareto frontier~\cite{zietlow2022leveling}, we propose a novel metric for evaluating the \textbf{Fairness-Accuracy Trade-off Efficiency (FATE)}, urging researchers to pay attention to the performance and fairness simultaneously when building prediction models. We evaluate the proposed method based on its application to mitigating unfairness in dermatology diagnosis.

To sum up, our contributions are as follows:
\begin{enumerate}
    \item A novel framework is proposed for unfairness mitigation by replacing normalization layers in backbones with FairAdaBN;
    \item A loss function is proposed to minimize statistical parity between subgroups for improving fairness;
    \item A new metric is derived to evaluate the model's fairness-performance trade-off efficiency. Our proposed FairAdaBN has the highest $\text{FATE}_{EOpp0}$ (48.79$\times10^{-2}$), which doubles the highest among other unfairness mitigation methods (Ind, 22.63 $\times10^{-2}$).
    \item Experiments on two dermatological disease datasets and three backbones demonstrate the superiority of our proposed FairAdaBN framework in terms of high performance and great portability.
\end{enumerate}

\section{Related Work}

According to~\cite{deho2022existing}, unfairness mitigation can be categorized into pre-processing, in-processing, and post-processing based on the instruction stage. 


\noindent{\textbf{{Pre-Processing.}}}
Pre-processing methods focus on the quality of the training set, by organizing fair datasets via datasets combination~\cite{seyyed2020chexclusion}, using generative adversarial networks~\cite{joshi2021ai} or sketching model~\cite{yao2022improving} to generate extra images, or directly resampling the train set~\cite{puyol2021fairness,zhang2022improving}. However, most methods in this category need huge effort due to the preciousness of medical data.

\noindent{\textbf{Post-Processing.}}
Although calibration has been widely used in unfairness mitigation in machine learning tasks, medical applications prefer to use pruning strategies. For example, Wu~\textit{et. al}~\cite{wu2022fairprune} mitigate unfairness by pruning a pre-trained diagnosis model considering the difference of feature importance between subgroups. However, their method needs extra time except for training a precise classification model, while our FairAdaBN is a one-step method.

\noindent{\textbf{In-Processing.}}
In-processing methods mainly consist of two folds. Some studies mitigate unfairness by directly adding fairness constraints to the cost functions~\cite{zhang2022improving}, which often leads to overfitting. 
Another category of research mitigates unfairness by designing complex network architectures like adversarial network~\cite{zhao2020training,li2021estimating} or representation learning~\cite{deng2023fairness}. This family of methods relies heavily on the accuracy of sensitive attribute classifiers in the adversarial branch,
leads to bigger models and cannot make full use of pre-trained weights.
While our method does not increase the number of parameters significantly and can be applied to several common backbones for dermatology diagnosis.

\section{FairAdaBN}

\noindent{\textbf{Problem Definition.}}
We assume a medical imaging dataset $D = \{d_1, d_2, ..., d_N\}$ with $N$ samples , the $i$-th sample $d_i$ consists of input image $X_i$, sensitive attributes $A_i$ and classification ground truth label $Y_i$. i.e. $d_i = \{X_i, A_i, Y_i\}$. $A$ is a binary variable (e.g., skin tone, gender), which splits the dataset into the unprivileged group, $D_{A=0}$, which has a lower average performance than the overall performance, and the privileged group, $D_{A=1}$, which has a higher average performance than the overall performance. 
Using accuracy as the performance metric for example, for a neural network $f_{\theta}(\cdot)$, our goal is to minimize the accuracy gap between $D_{A=0}$ and $D_{A=1}$ by finding a proper $\hat{\theta}$.

\begin{equation}
    \hat{\theta} = \arg\min_{\theta} \left\lVert \mathbb{E}_{\{X_i, Y_i\} \sim D_{A=1}}\mathbb{I}(f_{\theta}(X_i) = Y_i) - \mathbb{E}_{\{X_i, Y_i\} \sim D_{A=0}}\mathbb{I}(f_{\theta}(X_i) = Y_i) \right\rVert 
\end{equation}


\begin{figure}[ht]
    \centering
    \includegraphics[width=1.0\textwidth]{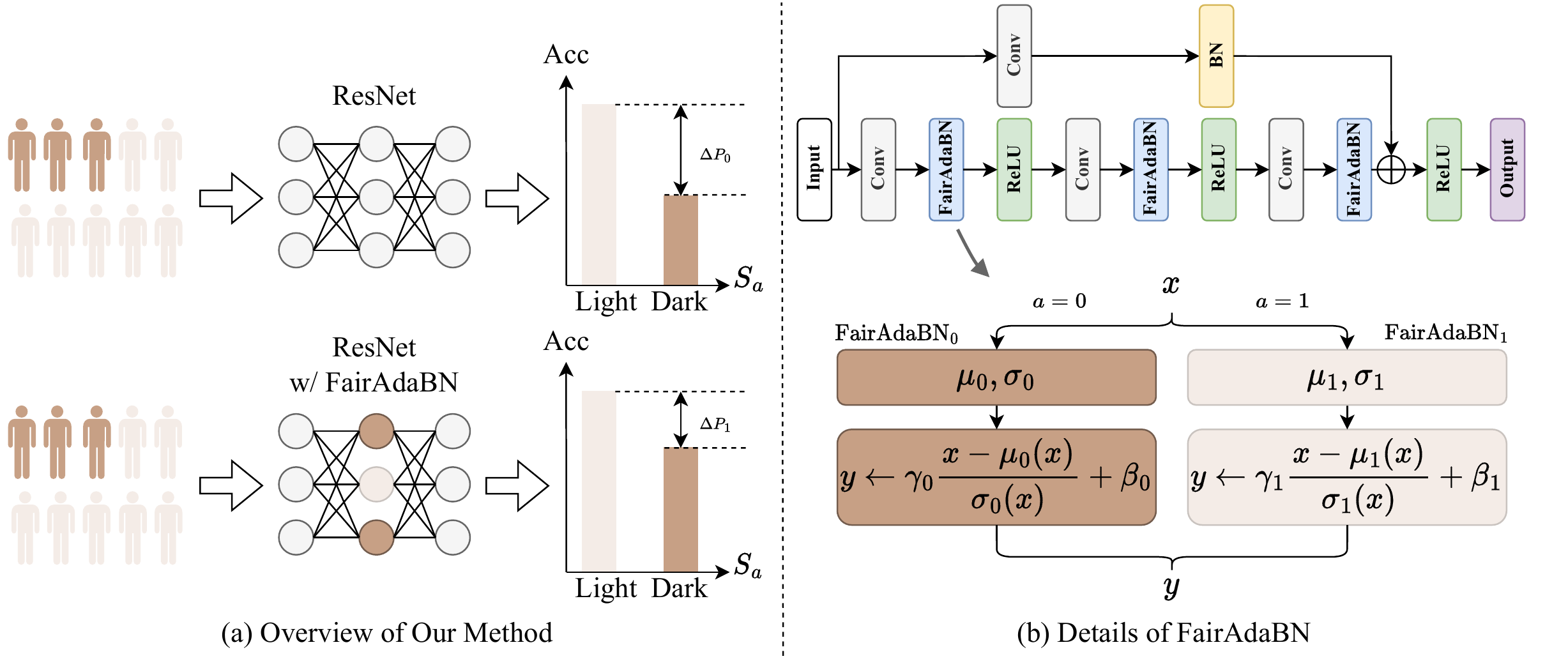}
    \caption{Overview of Our Method. (a) Compared to ResNet (Top), ResNet w/ AdaBN (Bottom) has a smaller performance disparity ($\Delta p$) between \textit{light} samples and \textit{dark} samples. (b) Details of FairAdaBN (use Residual block as an example).}
    \label{fig:overview}
\end{figure}


In this paper, we propose FairAdaBN, which replaces normalization layers in vanilla models with adaptive batch normalization layers, while sharing other layers between subgroups. The overview of our method is shown in Fig.~\ref{fig:overview}.

Batch normalization (BN) is a ubiquitous network layer that normalizes mini-batch features using statistics~\cite{ioffe2015batch}. Let $x \in \mathbb{R}^{C\times W\times H}$ denote a given layer's output feature map, where $C, W, H$ is the number of channels, width, and height of the feature map. The BN function is defined as:

\begin{equation}
    \text{BN}(x) = \gamma \cdot \frac{x - \mu(x)}{\sigma(x)} + \beta,
\end{equation}
where $\mu(x), \sigma(x)$ is the mean and standard deviation of the feature map computed in the mini-batch, $\beta$ and $\gamma$ denotes the learnable affine parameters.

We implant the attribute awareness into BN, named FairAdaBN, by parallelizing multiple normalization blocks that are carefully designed for each subgroup. Specifically, for subgroup $D_{A=a}$, its adaptive affine parameter $\gamma_{a}$ and $\beta_{a}$ are learnt by samples in $D_{A=a}$. Thus, the adaptive BN function for subgroup $D_{A=a}$ is given by Eq.~\ref{AdaBN}.

\begin{equation}
    \text{FairAdaBN}_a(x) = \gamma_a \cdot \frac{x - \mu_{a}(x)}{\sigma_{a}(x)} + \beta_a,
    \label{AdaBN}
\end{equation}
where $a$ is the index of the sensitive attribute corresponding to the current input image, $\mu_{\alpha}, \sigma_{\alpha}$ are computed across subgroups independently.

The FairAdaBN acquires subgroup-specific knowledge by learning the affine parameter $\gamma$ and $\beta$. Therefore, the feature maps of subgroups can be aligned and the unfair representation between privileged and unprivileged groups can be mitigated.
By applying FairAdaBN on vanilla backbones, the network can learn subgroup-agnostic feature representations by the sharing parameters of convolution layers, and subgroup-specific feature representations using respective BN parameters, resulting in lower fairness criteria. 
The detailed structure of FairAdaBN is shown in Fig.~\ref{fig:overview}, we display the minimum unit of ResNet for simplification. Note that the normalization layer in the residual branch is not changed for faster convergence.


In this paper, we aim to retain skin lesion classification accuracy and improve model fairness simultaneously. The loss function consists of two parts: (i) the cross-entropy loss, $L_{CE}$, constraining the prediction precision, and (2) the statistical disparity loss $L_{SD}$ as in Eq.~\ref{loss_sd}, aiming to minimize the difference of prediction probability between subgroups and give extra limits on fairness.
\begin{equation}
    L_{SD} = \sum_{y=1}^{N_{cg}}\left\lVert \mathbb{E}_{X_i \sim D_{A=0}} \mathbb{I}(f_{\theta}(X_i)=y) - \mathbb{E}_{X_i \sim D_{A=1}} \mathbb{I}(f_{\theta}(X_i)=y) \right\rVert^2,
    \label{loss_sd}
\end{equation}
where $N_{cg}$ means the number of classification categories. 

The overall loss function is given by the sum of the two parts, with a hyper-parameter $\alpha$ to adjust the degree of constraint on fairness.
    $L = L_{CE} + \alpha \cdot L_{SD}$.

\section{Experiments and Results}
\subsection{Evaluation Metrics}
Lots of fairness criteria are proposed including statistical parity~\cite{dwork2012fairness}, equalized odds~\cite{hardt2016equality}, equal opportunity~\cite{hardt2016equality}, counterfactual fairness~\cite{kusner2017counterfactual}, etc. In this paper, we use equal opportunity and equalized odds as fairness criteria. For equal opportunity, we split it into $EOpp0$ and $EOpp1$ considering the ground truth label.

\begin{equation}
     \text{EOpp0} = \lvert P(\hat{Y}=0 \mid Y=0, A=1) - P(\hat{Y}=0 \mid Y=0, A=0) \rvert
\end{equation}

\begin{equation}
    \text{EOpp1} = \lvert P(\hat{Y}=1 \mid Y=1, A=1) - P(\hat{Y}=1 \mid Y=1, A=0) \rvert
\end{equation}

\begin{equation}
    \text{EOdd} = \lvert P(\hat{Y}=1\mid Y=y, A=1)-P(\hat{Y}=1\mid Y=y, A=0)\rvert, y\in\{0,1\} 
\end{equation}

However, these metrics only evaluate the level of fairness while do not consider the trade-off between fairness and accuracy. 
Therefore, inspired by~\cite{dhar2021pass}, we propose FATE, a metric that evaluates the balance between normalized improvement of fairness and normalized drop of accuracy. The formulas of FATE on different fairness criteria are shown below:

\begin{equation}
    \text{FATE}_{FC} = \frac{\text{ACC}_m - \text{ACC}_b}{\text{ACC}_b} - \lambda \frac{\text{FC}_m - \text{FC}_b}{\text{FC}_b},
\end{equation}
where $FC$ can be one of $EOpp0, EOpp1, EOdd$. $ACC$ denotes accuracy. The subscript $m$ and $b$ denote the mitigation model and baseline model, respectively. $\lambda$ is a weighting factor that adjusts the requirements for fairness pre-defined by the user considering the real application, here we define $\lambda=1.0$ for simplification. A model obtains a higher FATE if it mitigates unfairness and maintains accuracy. Note that FATE should be combined with utility metrics and fairness metrics, rather than independently.

\subsection{Dataset and Network Configuration}

We use two well-known dermatology datasets to evaluate the proposed method. 
The Fitzpatrick-17k dataset\cite{groh2022towards} contains 16,577 dermatology images in 9 diagnostic categories. The skin tone is labeled with Fitzpatrick's skin phenotype. In this paper, we regard Skin Type I to III as \textit{light}, and Skin Type IV to VI as \textit{dark} for simplicity, resulting in a ratio of $\text{dark}:\text{light} \approx 3:7$. 
The ISIC 2019 dataset\cite{tschandl2018ham10000,codella2018skin,combalia2019bcn20000} contains 25,331 images among 9 different diagnostic categories. We use gender as the sensitive attribute, where $\text{female}:\text{male} \approx 4.5:5.5$. Based on subgroup analysis, \textit{dark} and \textit{female} are treated as the privileged group, and \textit{light} and \textit{male} are treated as the unprivileged group.

We randomly split the dataset into train, validation, and test with a ratio of 6:2:2. The models are trained for 600 epochs and the model with the highest validation accuracy is selected for testing.
The images are resized or cropped to 128 $\times$ 128 for both datasets. Random flipping and random rotation are used for data augmentation. 
The experiments are carried out on 8 $\times$ NVIDIA 3090 GPUs, implemented on PyTorch, and are repeated 3 times. Pre-trained weights from ImageNet are used for all models. The networks are trained using AdamW optimizer with weight decay. The batch size and learning rate are set as 128 and 1e-4, respectively. The hyper-parameter $\alpha = 1.0$.

\subsection{Results}

We compare FairAdaBN with Vanilla (ResNet-152), Resampling~\cite{puyol2021fairness}, Ind (independently trained models for each subgroup)~\cite{puyol2021fairness}, GroupDRO~\cite{sagawa2019distributionally}, EnD~\cite{tartaglione2021end}, and CFair~\cite{zhao2019conditional}, which are commonly used for unfairness mitigation.

\noindent{\textbf{Results on Fitzpatrick-17k Dataset.}} Table~\ref{tab:exp_fitz} shows the result of these seven methods on Fitzpatrick-17k dataset. 
Compared to the Vanilla model, Resampling has a comparable utility, but cannot improve fairness.
FairAdaBN achieves the lowest unfairness with only a small drop in accuracy. 
Besides, FairAdaBN has the highest FATE on all fairness criteria. 
This is because Ind does not share common information between subgroups, and only part of the dataset is used for training. 
GroupDRO and EnD rely on the discrimination of features from different subgroups, which is indistinguishable for this task.
CFair is more efficient on balanced datasets, while the ratio between $light$ and $dark$ is skewed.

\noindent{\textbf{Results on ISIC 2019 Dataset.}} Table~\ref{tab:exp_fitz} shows the results on ISIC 2019 dataset. FairAdaBN is the fairest method among the seven methods. Resampling improves fairness sightly but does not outperform ours. GroupDRO mitigates EOpp0 while increasing unfairness on Eopp1 and Eodd. Ind and CFair cannot mitigate unfairness in ISIC 2019 dataset and EnD increases unfairness on EOpp0. 

\linespread{1.1}
\begin{table}[ht]
    \centering
    \caption{Result on Fitzpatrick-17k and ISIC 2019 Dataset ($\text{Mean}^{\text{Std}} \times 10^{-2}$). \textbf{Best} and \emph{Second-best} are highlighted.}
    \label{tab:exp_fitz}
    \resizebox{\textwidth}{!}{
    \begin{tabular}{l|rrrr|rrr|rrr}
        \hline
        \hline
        \multicolumn{11}{c}{\textbf{Fitzpartrick-17k Dataset}} \\
        \hline
        Method & Accuracy$\uparrow$& Precision$\uparrow$ & Recall$\uparrow$& F1$\uparrow$  & EOpp0$\downarrow$ & EOpp1$\downarrow$ & Eodd$\downarrow$ & $\text{E}_{0}\uparrow$ & $\text{E}_{1}\uparrow$ & $\text{E}_{2}\uparrow$\\
        \hline
        Vanilla & {$87.53^{0.14}$} & \bm{$79.60^{0.33}$} & \bm{$80.22^{0.19}$} & \bm{$78.41^{0.15}$} & $1.00^{0.30}$ & $10.40^{1.43}$ & $10.54^{0.98}$ & / & / & / \\
        \hline
        Resampling~\cite{puyol2021fairness}$\dagger$ & \uline{$87.73^{0.27}$} & \uline{$79.21^{0.40}$} & \uline{$80.01^{0.35}$} & \uline{$78.27^{0.42}$} & {$1.11^{0.26}$} & $10.43^{1.91}$ & $10.78^{2.06}$ & {$-10.86$} & $-0.03$ & $-2.05$\\
        \hline        
        Ind~\cite{puyol2021fairness}$\dagger$ & $86.33^{0.12}$ & $76.11^{0.38}$ & $77.48^{0.18}$ & $75.20^{0.09}$ & \uline{$0.78^{0.33}$} & $10.13^{0.51}$ & $9.72^{0.94}$ & \uline{$20.63$} & $1.23$ & $6.41$\\
        \hline
        GroupDRO~\cite{sagawa2019distributionally}$\dagger$ & $86.62^{0.19}$ & $77.21^{0.62}$ & $78.29^{0.52}$ & $76.56^{0.56}$ & $0.94^{0.34}$ & \uline{$8.04^{0.90}$} & \uline{$8.23^{1.25}$} & $5.07$ & \uline{$21.66$} & \uline{$20.91$}\\
        \hline
        EnD~\cite{tartaglione2021end}$\dagger$& $86.80^{0.52}$ & $77.32^{0.60}$ & $78.58^{0.53}$ & $76.90^{0.66}$ & $1.22^{0.31}$ & {$9.01^{1.60}$} & {$9.20^{1.59}$} & $-22.83$ & {$12.53$} & {$11.88$}\\
        \hline
        CFair~\cite{zhao2019conditional}$\dagger$& \bm{$87.91^{0.35}$} & {$78.62^{0.49}$} & {$79.73^{0.37}$} & {$78.12^{0.38}$} & $0.93^{0.28}$ & $9.83^{1.65}$ & $10.17^{1.57}$ & $10.03$ & $12.15$ & $10.09$\\
        \hline
        FairAdaBN & $84.72^{0.40}$ & $74.43^{0.22}$ & $75.74^{0.33}$ & $73.31^{0.48}$ & \bm{$0.48^{0.09}$} & \bm{$7.67^{3.86}$} & \bm{$7.73^{3.95}$} & \bm{$48.79$} & \bm{$23.04$} & \bm{$23.45$}\\
        \hline
        \hline
        \multicolumn{11}{c}{\textbf{ISIC 2019 Dataset}} \\
        \hline
        Method & Accuracy$\uparrow$& Precision$\uparrow$ & Recall$\uparrow$& F1$\uparrow$  & EOpp0$\downarrow$ & EOpp1$\downarrow$ & Eodd$\downarrow$ & $\text{E}_{0}\uparrow$ & $\text{E}_{1}\uparrow$ & $\text{E}_{2}\uparrow$\\
        \hline
        Vanilla & \uline{$92.52^{0.12}$} & \uline{$82.64^{0.31}$} & \uline{$82.94^{0.36}$} & \uline{$82.60^{0.32}$} & {$0.85^{0.12}$} & $6.12^{1.83}$ & $6.02^{1.66}$ & / & / & / \\
        \hline
        Resampling~\cite{puyol2021fairness}$\dagger$ & \bm{$92.81^{0.28}$} & \bm{$83.15^{0.50}$} & \bm{$83.42^{0.51}$} & \bm{$83.12^{0.52}$} & {$0.86^{0.15}$} & $5.65^{2.83}$ & $5.76^{2.78}$ & -0.80 & -2.48 & -5.49 \\
        \hline
        Ind~\cite{puyol2021fairness}$\dagger$& {$92.43^{0.11}$} & {$82.16^{0.15}$} & {$82.46^{0.12}$} & {$82.11^{0.08}$} & $0.85^{0.11}$ & $7.04^{0.96}$ & $7.37^{0.77}$ & {$-0.10$} & $-15.13$ & $-22.52$\\
        \hline
        GroupDRO~\cite{sagawa2019distributionally}$\dagger$ & $91.86^{0.22}$ & $81.30^{0.52}$ & $81.44^{0.47}$ & $81.17^{0.50}$ & \uline{$0.82^{0.12}$} & {$6.78^{3.20}$} & {$6.62^{3.21}$} & \uline{$2.41$} & {$-22.99$} & {$-22.01$}\\
        \hline
        EnD~\cite{tartaglione2021end}$\dagger$& $92.13^{0.08}$ & $81.42^{0.48}$ & $81.64^{0.35}$ & $81.36^{0.38}$ & $0.98^{0.09}$ & \uline{$5.18^{0.99}$} & \uline{$5.10^{1.06}$} & $-15.72$ & \uline{$14.94$} & \uline{$14.86$}\\
        \hline
        CFair~\cite{zhao2019conditional}$\dagger$& $87.39^{0.77}$ & $72.39^{2.67}$ & $72.60^{2.22}$ & $71.28^{2.12}$ & $2.83^{1.09}$ & $9.21^{3.53}$ & $10.80^{4.15}$ & $-238.49$ & $-56.03$ & $-84.95$\\
        \hline
        FairAdaBN & $89.11^{0.09}$ & $74.24^{0.13}$ & $74.79^{0.18}$ & $74.18^{0.14}$ & \bm{$0.69^{0.07}$} & \bm{$4.85^{2.50}$} & \bm{$4.76^{2.73}$} & \bm{$15.14$} & \bm{$17.07$} & \bm{$17.24$}\\
        \hline
    \end{tabular}
    }
    \begin{tablenotes}
        \item * $E_0, E_1, E_2$ denotes $\text{FATE}_{EOpp0}, \text{FATE}_{EOpp1}, \text{FATE}_{EOdd}$, respectively.
        \item $\dagger$  Private implementation.
    \end{tablenotes}
\end{table}
\linespread{1.0}

\begin{figure}[ht]
    \centering
    \includegraphics[width=1.0\textwidth]{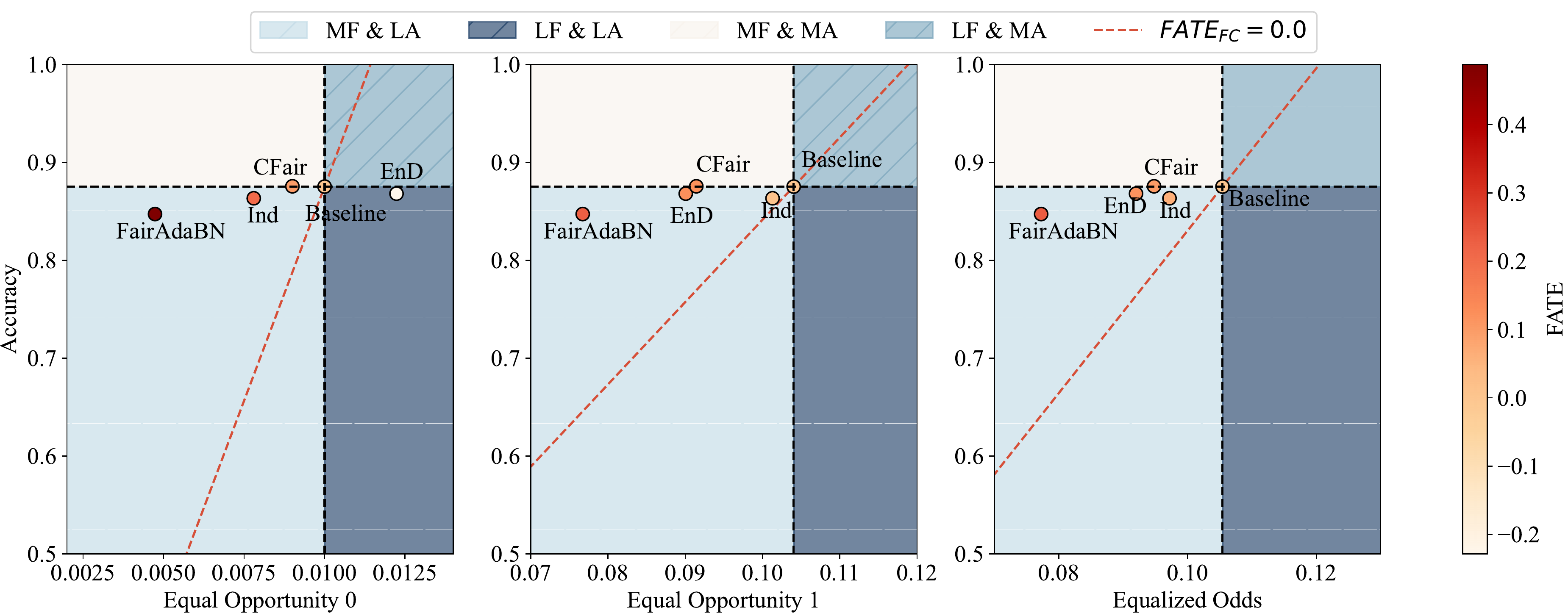}
    \caption{FATE on different fairness criteria. The data point of baseline (fairness, accuracy) splits the space into four parts: MF: Fairer; MA: More Accurate; LF: Less Fair; LA: Less Accurate. Points on the left of the line have positive FATE, while points on the right of the line have negative FATE. }
    \label{fig:fate_result}
\end{figure}

\noindent{\textbf{The FATE metric.}} Fig.~\ref{fig:fate_result} shows the values of FATE.
According to~\cite{creager2019flexibly}, the closer the curve is to the top left corner, the smaller the fairness-accuracy trade-off it has. The figure demonstrates that FATE has the same trend as this argument. We prefer an algorithm that obtains a higher FATE since a higher FATE denotes higher unfairness mitigation and a low drop in utility, and a negative FATE denotes that the mitigation model cannot decrease unfairness while reserving enough accuracy (not beneficial).

\linespread{1.1}
\begin{table}[ht]
    \centering
    \caption{Ablation Study ($\text{Mean}^{\text{Std}} \times 10^{-2}$). \textbf{Best} in each group are highlighted.}
    \label{tab:Ablation Study}
    \resizebox{\textwidth}{!}{
    \begin{tabular}{l|rrrr|rrr|rrr}
    \hline
    Method & Accuracy$\uparrow$& Precision$\uparrow$ & Recall$\uparrow$& F1$\uparrow$  & EOpp0$\downarrow$ & EOpp1$\downarrow$ & Eodd$\downarrow$ & $\text{E}_{0}\uparrow$ & $\text{E}_{1}\uparrow$ & $\text{E}_{2}\uparrow$\\
    \hline
    VGG & \bm{$88.11^{0.51}$} & \bm{$79.18^{0.56}$} &\bm{$80.07^{0.49}$} & \bm{$78.55^{0.56}$} & $1.42^{0.25}$ & $10.64^{2.15}$ & $11.78^{2.34}$ & / & / & / \\
    \hline
    VGG + FairAdaBN & $83.55^{0.24}$ & $69.73^{0.83}$ & $72.09^{0.41}$ & $70.15^{0.69}$ & \bm{$1.09^{0.04}$} & \bm{$10.58^{1.80}$} & \bm{$10.48^{1.97}$} & 18.06 & -4.61 & 5.86  \\
    \hline
    \hline
    DenseNet & \bm{$87.32^{0.06}$} & \bm{$78.12^{0.52}$} & \bm{$79.08^{0.38}$} & \bm{$77.37^{0.24}$} & \bm{$1.18^{0.37}$} & $10.96^{1.34}$ & $11.47^{1.16}$ & / & / & / \\
    \hline
    DenseNet + FairAdaBN & $80.40^{0.23}$ & $65.32^{0.57}$ & $69.42^{0.40}$ & $65.25^{0.60}$ & $1.43^{0.79}$ & \bm{$7.70^{1.06}$} & \bm{$8.30^{1.58}$} & -29.11 & 21.82 & 19.71 \\   
    \hline
    \hline
    ResNet & \bm{$87.53^{0.14}$} & \bm{$79.60^{0.33}$} & \bm{$80.22^{0.19}$} & \bm{$78.41^{0.15}$} & $1.00^{0.30}$ & $10.40^{1.43}$ & $10.54^{0.98}$ & / & / & /  \\
    \hline
    Ours w/o $L_{SD}$ & $87.18^{0.50}$ & $78.50^{0.75}$ & $79.24^{0.68}$ & $77.40^{0.71}$ & $1.07^{0.16}$ & $9.33^{0.23}$ & $9.91^{0.29}$ & -7.87 & 9.88 & 5.55  \\
    \hline
    Ours w/o FairAdaBN & $85.02^{0.03}$ & $73.76^{0.11}$ & $75.67^{0.05}$ & $73.63^{0.16}$ & $1.39^{0.45}$ & $15.30^{1.91}$ & $15.05^{1.37}$ & 42.15 & -49.94 & -45.62 \\
    \hline
    Ours ($\alpha = 0.1$) & $84.82^{0.79}$ & $73.44^{1.11}$ & $75.15^{0.98}$ & $73.17^{0.95}$ & $1.26^{0.18}$ & $13.39^{2.98}$ & $12.76^{3.28}$ & -29.10 & -31.85 &  -24.16 \\
    \hline 
    Ours ($\alpha = 1.0$) & $84.72^{0.40}$ & $74.43^{0.22}$ & $75.74^{0.33}$ & $73.31^{0.48}$  & \bm{$0.48^{0.09}$} & \bm{$7.67^{3.86}$} & \bm{$7.73^{3.95}$} & \bm{$48.79$} & \bm{$23.04$} & \bm{$23.45$}\\
    \hline 
    Ours ($\alpha = 2.0$) & $84.57^{0.38}$ & $74.26^{0.22}$ & $75.40^{0.11}$ & $72.91^{0.87}$ & $1.10^{0.60}$ & $8.53^{2.79}$ & $8.40^{2.75}$ & -13.38 & 14.60 & 16.92 \\   
    \hline
    \end{tabular}
    }
\end{table}
\linespread{1.0}

\noindent\textbf{{Limitation.}}
Compared with other methods, FairAdaBN needs to use sensitive attributes in the test stage, which is unnecessary for EnD and CFair. Although this might be easy to acquire in real applications, improvements could be done to solve this problem.

\subsection{Ablation Study}

\textbf{Different backbones.} Firstly, we test FairAdaBN's compatibility on different backbones, by applying FairAdaBN on VGG-19-BN and DenseNet-121. Note that the first and last BN in DenseNet are not changed. The result is shown in Table~\ref{tab:Ablation Study}. 
The experiments are carried out on Fitzpatrick-17k dataset.
The result shows that our FairAdaBN is also effective on these two backbones, except $Eopp_0$ when using DenseNet-121, showing well model compatibility. However, we also observe a larger drop in model precision compared with the baseline, which needs to be taken into consideration in future work.

\noindent\textbf{Different loss terms.} We train ResNet by only replacing BNs with FairAdaBNs (the second row of the last part), and ResNet adding $L_{SD}$ on the total loss (the third row of the last part). 
The effectiveness of AdaBN is illustrated by comparing the first and second rows of the last part in Table~\ref{tab:Ablation Study}. By replacing BNs with FairAdaBN, ResNet can normalize subgroup feature maps using specific affine parameters, which reduce $Eopp_1$ and $Eodd$ by $1.07\times10^{-2}$ and $0.63\times10^{-2}$, respectively.
Comparing the second and fourth row of the last part in Table~\ref{tab:Ablation Study}, we find that by adding $L_{SD}$, $Eopp_0$ decreases significantly, from $1.07 \times 10^{-2}$ to $0.48 \times 10^{-2}$. 
Besides, although adding $L_{SD}$ on ResNet alone increases fairness criteria unexpectedly, fairness criteria decrease when using FairAdaBN and $L_{SD}$ simultaneously. The reason could be the potential connection between FairAdaBN and $L_{SD}$, due to the similar form dealing with subgroups.

\noindent\textbf{Hyper-parameter $\alpha$.} Our experiments show that $\alpha=1.0$ has the best fairness scores and FATE compared to $\alpha=0.1$ and $\alpha=2.0$. Therefore we select $\alpha=1.0$ as our final setting.

\section{Conclusion}
We propose FairAdaBN, a simple but effective framework for unfairness mitigation in dermatological disease classification. Extensive experiments illustrate that the proposed framework can mitigate unfairness compared to models without fair constraints, and has a higher fairness-accuracy trade-off efficiency compared with other unfairness mitigation methods. By plugging FairAdaBN into several backbones, its generalization ability is proved. However, the current study only evaluates the effectiveness of FairAdaBN on dermatology datasets, and its generalization ability on other datasets (chest X-Ray, brain MRI) or tasks (segmentation, detection), where unfairness issues also exist, needs to be evaluated in the future. We also plan to explore the unfairness mitigation effectiveness for other universal models~\cite{zhou2021review}.

%
%
%
\bibliographystyle{splncs04}
\bibliography{paper1119}

\begin{thebibliography}{10}
\providecommand{\url}[1]{\texttt{#1}}
\providecommand{\urlprefix}{URL }
\providecommand{\doi}[1]{https://doi.org/#1}

\bibitem{codella2018skin}
Codella, N.C., Gutman, D., Celebi, M.E., Helba, B., Marchetti, M.A., Dusza,
  S.W., Kalloo, A., Liopyris, K., Mishra, N., Kittler, H., et~al.: Skin lesion
  analysis toward melanoma detection: A challenge at the 2017 international
  symposium on biomedical imaging ({ISBI}), hosted by the international skin
  imaging collaboration ({ISIC}). In: 2018 IEEE 15th international symposium on
  biomedical imaging (ISBI 2018). pp. 168--172. IEEE (2018)

\bibitem{combalia2019bcn20000}
Combalia, M., Codella, N.C., Rotemberg, V., Helba, B., Vilaplana, V., Reiter,
  O., Carrera, C., Barreiro, A., Halpern, A.C., Puig, S., et~al.: {BCN20000}:
  Dermoscopic lesions in the wild. arXiv preprint arXiv:1908.02288  (2019)

\bibitem{creager2019flexibly}
Creager, E., Madras, D., Jacobsen, J.H., Weis, M., Swersky, K., Pitassi, T.,
  Zemel, R.: Flexibly fair representation learning by disentanglement. In:
  International conference on machine learning. pp. 1436--1445. PMLR (2019)

\bibitem{deho2022existing}
Deho, O.B., Zhan, C., Li, J., Liu, J., Liu, L., Duy~Le, T.: How do the existing
  fairness metrics and unfairness mitigation algorithms contribute to ethical
  learning analytics? British Journal of Educational Technology  (2022)

\bibitem{deng2023fairness}
Deng, W., Zhong, Y., Dou, Q., Li, X.: On fairness of medical image
  classification with multiple sensitive attributes via learning orthogonal
  representations. arXiv preprint arXiv:2301.01481  (2023)

\bibitem{dhar2021pass}
Dhar, P., Gleason, J., Roy, A., Castillo, C.D., Chellappa, R.: {PASS}:
  Protected attribute suppression system for mitigating bias in face
  recognition. In: Proceedings of the IEEE/CVF International Conference on
  Computer Vision. pp. 15087--15096 (2021)

\bibitem{dwork2012fairness}
Dwork, C., Hardt, M., Pitassi, T., Reingold, O., Zemel, R.: Fairness through
  awareness. In: Proceedings of the 3rd innovations in theoretical computer
  science conference. pp. 214--226 (2012)

\bibitem{groh2022towards}
Groh, M., Harris, C., Daneshjou, R., Badri, O., Koochek, A.: Towards
  transparency in dermatology image datasets with skin tone annotations by
  experts, crowds, and an algorithm. arXiv preprint arXiv:2207.02942  (2022)

\bibitem{hardt2016equality}
Hardt, M., Price, E., Srebro, N.: Equality of opportunity in supervised
  learning. Advances in neural information processing systems  \textbf{29}
  (2016)

\bibitem{ioffe2015batch}
Ioffe, S., Szegedy, C.: Batch normalization: Accelerating deep network training
  by reducing internal covariate shift. In: International conference on machine
  learning. pp. 448--456. pmlr (2015)

\bibitem{joshi2021ai}
Joshi, N., Burlina, P.: {AI} fairness via domain adaptation. arXiv preprint
  arXiv:2104.01109  (2021)

\bibitem{kinyanjui2020fairness}
Kinyanjui, N.M., Odonga, T., Cintas, C., Codella, N.C., Panda, R., Sattigeri,
  P., Varshney, K.R.: Fairness of classifiers across skin tones in dermatology.
  In: International Conference on Medical Image Computing and Computer-Assisted
  Intervention. pp. 320--329. Springer (2020)

\bibitem{kusner2017counterfactual}
Kusner, M.J., Loftus, J., Russell, C., Silva, R.: Counterfactual fairness.
  Advances in neural information processing systems  \textbf{30} (2017)

\bibitem{li2021estimating}
Li, X., Cui, Z., Wu, Y., Gu, L., Harada, T.: Estimating and improving fairness
  with adversarial learning. arXiv preprint arXiv:2103.04243  (2021)

\bibitem{lu2021evaluating}
Lu, C., Lemay, A., Hoebel, K., Kalpathy-Cramer, J.: Evaluating subgroup
  disparity using epistemic uncertainty in mammography. arXiv preprint
  arXiv:2107.02716  (2021)

\bibitem{narayanan2018translation}
Narayanan, A.: Translation tutorial: 21 fairness definitions and their
  politics. In: Proc. Conf. Fairness Accountability Transp., New York, USA.
  vol.~1170, p.~3 (2018)

\bibitem{petersen2022feature}
Petersen, E., Feragen, A., Zemsch, L.d.C., Henriksen, A., Christensen, O.E.W.,
  Ganz, M.: Feature robustness and sex differences in medical imaging: a case
  study in mri-based alzheimer's disease detection. arXiv preprint
  arXiv:2204.01737  (2022)

\bibitem{puyol2021fairness}
Puyol-Ant{\'o}n, E., Ruijsink, B., Piechnik, S.K., Neubauer, S., Petersen,
  S.E., Razavi, R., King, A.P.: Fairness in cardiac {MR} image analysis: An
  investigation of bias due to data imbalance in deep learning based
  segmentation. In: International Conference on Medical Image Computing and
  Computer-Assisted Intervention. pp. 413--423. Springer (2021)

\bibitem{sagawa2019distributionally}
Sagawa, S., Koh, P.W., Hashimoto, T.B., Liang, P.: Distributionally robust
  neural networks for group shifts: On the importance of regularization for
  worst-case generalization. arXiv preprint arXiv:1911.08731  (2019)

\bibitem{sarhan2020fairness}
Sarhan, M.H., Navab, N., Eslami, A., Albarqouni, S.: On the fairness of
  privacy-preserving representations in medical applications. In: Domain
  Adaptation and Representation Transfer, and Distributed and Collaborative
  Learning, pp. 140--149. Springer (2020)

\bibitem{seyyed2020chexclusion}
Seyyed-Kalantari, L., Liu, G., McDermott, M., Chen, I.Y., Ghassemi, M.:
  Chexclusion: Fairness gaps in deep chest {X}-ray classifiers. In:
  BIOCOMPUTING 2021: Proceedings of the Pacific Symposium. pp. 232--243. World
  Scientific (2020)

\bibitem{suriyakumar2021chasing}
Suriyakumar, V.M., Papernot, N., Goldenberg, A., Ghassemi, M.: Chasing your
  long tails: Differentially private prediction in health care settings. In:
  Proceedings of the 2021 ACM Conference on Fairness, Accountability, and
  Transparency. pp. 723--734 (2021)

\bibitem{tartaglione2021end}
Tartaglione, E., Barbano, C.A., Grangetto, M.: En{D}: Entangling and
  disentangling deep representations for bias correction. In: Proceedings of
  the IEEE/CVF conference on computer vision and pattern recognition. pp.
  13508--13517 (2021)

\bibitem{tschandl2018ham10000}
Tschandl, P., Rosendahl, C., Kittler, H.: The {HAM10000} dataset, a large
  collection of multi-source dermatoscopic images of common pigmented skin
  lesions. Scientific data  \textbf{5}(1), ~1--9 (2018)

\bibitem{wang2020mitigating}
Wang, M., Deng, W.: Mitigating bias in face recognition using skewness-aware
  reinforcement learning. In: Proceedings of the IEEE/CVF conference on
  computer vision and pattern recognition. pp. 9322--9331 (2020)

\bibitem{wu2022fairprune}
Wu, Y., Zeng, D., Xu, X., Shi, Y., Hu, J.: Fairprune: Achieving fairness
  through pruning for dermatological disease diagnosis. arXiv preprint
  arXiv:2203.02110  (2022)

\bibitem{yao2022improving}
Yao, R., Cui, Z., Li, X., Gu, L.: Improving fairness in image classification
  via sketching. arXiv preprint arXiv:2211.00168  (2022)

\bibitem{zhang2022improving}
Zhang, H., Dullerud, N., Roth, K., Oakden-Rayner, L., Pfohl, S., Ghassemi, M.:
  Improving the fairness of chest x-ray classifiers. In: Conference on Health,
  Inference, and Learning. pp. 204--233. PMLR (2022)

\bibitem{zhao2019conditional}
Zhao, H., Coston, A., Adel, T., Gordon, G.J.: Conditional learning of fair
  representations. arXiv preprint arXiv:1910.07162  (2019)

\bibitem{zhao2020training}
Zhao, Q., Adeli, E., Pohl, K.M.: Training confounder-free deep learning models
  for medical applications. Nature communications  \textbf{11}(1), ~1--9 (2020)

\bibitem{zhou2021review}
Zhou, S.K., Greenspan, H., Davatzikos, C., Duncan, J.S., Van~Ginneken, B.,
  Madabhushi, A., Prince, J.L., Rueckert, D., Summers, R.M.: A review of deep
  learning in medical imaging: Imaging traits, technology trends, case studies
  with progress highlights, and future promises. Proceedings of the IEEE
  (2021)

\bibitem{zietlow2022leveling}
Zietlow, D., Lohaus, M., Balakrishnan, G., Kleindessner, M., Locatello, F.,
  Sch{\"o}lkopf, B., Russell, C.: Leveling down in computer vision: Pareto
  inefficiencies in fair deep classifiers. In: Proceedings of the IEEE/CVF
  Conference on Computer Vision and Pattern Recognition. pp. 10410--10421
  (2022)

\end{thebibliography}

\end{document}